\documentclass[journal, letterpaper]{IEEEtran}

\usepackage{graphicx}

\usepackage{url}

\usepackage{amsmath}

\usepackage{textgreek}	
\usepackage{listings}
\usepackage{csvsimple}
\usepackage{longtable}
\usepackage{url}
\usepackage{hyperref}

\begin{document}

	\title{A geometric approach towards inverse kinematics of soft extensible pneumatic actuators intended for trajectory tracking}

\author{Mahboubeh~Keyvanara$^{1,*}$, 
        Arman~Goshtasbi$^{2}$, 
        and~Irene~A.~Kuling$^{1}$ 
\thanks{$^{1}$ Reshape Lab, Department of Mechanical Engineering, Dynamics and Control Group, Eindhoven University of Technology, 5600 MB Eindhoven, The Netherlands.}
\thanks{$^{2}$Soft Robotics Laboratory, Department of Biomechanical Engineering, Engineering Technology Faculty, University of Twente, 7522 NB, The Netherlands }
\thanks{$^*$Corressponding author: m.keyvanara@tue.nl}}
	\markboth{}{}
	\maketitle

\begin{abstract}
	Soft robots are interesting examples of hyper-redundancy in robotics, however, the nonlinear continuous dynamics of these robots and the use of hyper-elastic and visco-elastic materials makes modeling of these robots more complicated. This study presents a geometric Inverse Kinematic (IK) model for trajectory tracking of multi-segment extensible soft robots, where, each segment of the soft actuator is geometrically approximated with multiple rigid links connected with rotary and prismatic joints. Using optimization methods, the desired configuration variables of the soft actuator for the desired end-effector positions are obtained. Also, the redundancy of the robot is applied for second task applications, such as tip angle control. The model's performance is investigated through simulations, numerical benchmarks, and experimental validations and results show lower computational costs and higher accuracy compared to most existing methods. The method is easy to apply to multi segment soft robots, both in 2D and 3D. As a case study, a fully 3D-printed soft robot manipulator is tested using a control unit and the model predictions show good agreement with the experimental results.
\end{abstract}


\section{Introduction}

\IEEEPARstart{S}{olutions} to the IK of soft robots have been studied using different approaches. In analytical solutions, using the geometric and kinematics of the robot, an absolute solution for the required DoF is found. This solution is not trivial due to these robots’ nonlinear equations and hyper-redundancy. To overcome this, different models are proposed, the most famous one being the constant curvature (CC) approximation \cite{PCC1,walker_PCC,PCC_review}. 
Using this model, authors in \cite{Analytical_nazari} suggested an absolute geometrical solution for a single-segment inextensible continuum arm, showing it to be the most suitable method for a single-segment robot. Yet, for multi-segment soft robots, the combination of piecewise constant curvature (PCC) model and analytical solutions can lead to complex mathematics, high computation costs, and so many simplifications are required \cite{underwater}. For this reason, a less intricate yet accurate model is needed to overcome the complexity of the analytical solution. 

\begin{figure}[t]
    \centering
    \includegraphics[width=0.5\textwidth]{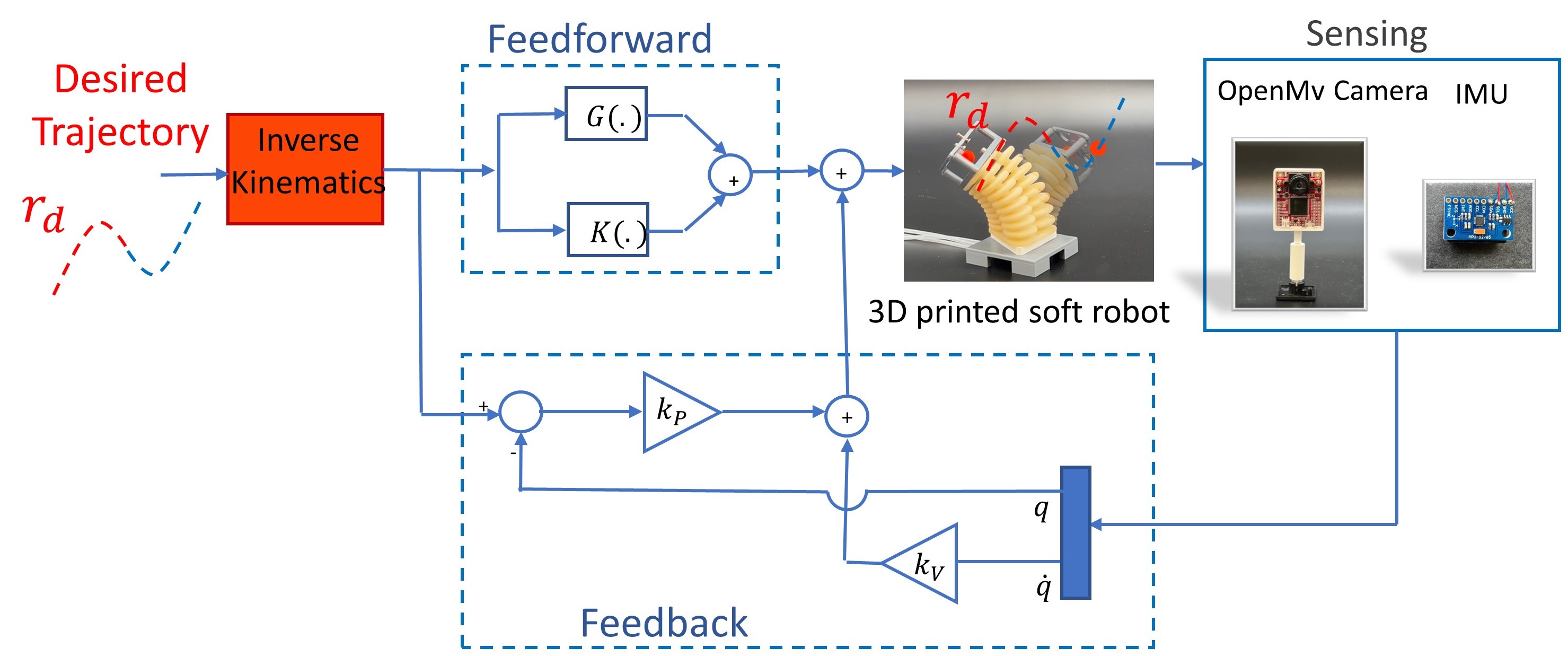}
   \caption{3D trajectory tracking of soft robots. A block diagram representing closed loop control of the robot on the desired trajectory. Parameters defined in this schematic are later defined in the paper.} 
    \label{fig:Intro}
\end{figure}

Numerical solutions are a more promising alternative for IK solutions where state variables are calculated approximately using iterative or optimization techniques.
In \cite{Fem_input}, deformations of a soft robot under different actuation loads are simulated using FEM.
The main benefit of the FEM model is its capability to solve the inverse kinematics of nonuniform shape robots and torsional robots (\cite{Geometric} \cite{Rus_FEM}), yet, the results are dependent very much on the robot itself.
This method is also computationally intensive for multi-link systems and real-time control. Only recently in \cite{SOFA_IK}, using visual servoing, the authors have performed closed-loop trajectory tracking of a soft robot using FEM.
In other numerical methods, such as curvature discretization, the nonlinear behavior of soft robots is approximated by multiple rigid links. In \cite{Mutlu2013AnEM}, authors suggest 16 rigid link approximation to model single segment polymer actuator. 
In \cite{Sarthak}, a more complicated shape of soft robots is approximated by rigid-link modeling. Here, the strain of the robot during actuation has not been taken into account and only bending is modeled and a solution for extensible soft robots is still missing. 
Other numerical methods, namely heuristics inverse kinematics algorithms, 
are famous for low computation costs and the possibility of solving large DoFs systems \cite{CCD_orginal}. 
\cite{CCD} proposed a new cyclic coordinate descent (CCD) algorithm for soft and redundant robots by solving the shortcoming of the standard CCD algorithms. 
Unfortunately, this method is limited to simulations only and has not been tested with real-time controllers and experimental setups.

Learning methods are also employed to overcome the complexity of the IK problem. Mostly, these techniques are combined with other well-known IK solutions to propose faster solutions \cite{Ge_fang}. 
For example, in \cite{Jacobian_learning}, authors combined the Jacobian inverse kinematics method with feed-forward neural networks to control not only the multi-link robot but also optimize energy usage. 
In learning approaches, similar to FEM approaches, inverse kinematics can be solved for nonuniform robots. However, a physical experimental setup or a sophisticated dynamic model is required for the data acquisition.


 Leveraging on this literature and with the aim of proposing a fast and accurate IK solution for extensible multi-segment 3D soft robots that can be used easily in experimental validations, we propose a geometric inverse kinematic model for 2D and 3D single and multi-segment extensible soft robots.
 In the proposed method, each robot segment is modeled with a CC approximation and discretized with multiple rigid links connected together with prismatic and rotary joints. The developed method is mathematically simple, can easily be applied to various types of soft actuators, and has negligible error with respect to the workspace. Increasing the DoF of the robot by adding segments to the robot is straightforward and does not significantly affect the method's performance.
 Using this geometric model, an optimization method is employed to find a configuration for the soft robot, which will result in the end-effector reaching the required position. The optimization method allows for working with the boundary conditions using redundancy of the robot for a second task, such as tip angle control and pressure control.
One advantage of the proposed IK model is its application diversity, as it is possible to employ this method with any desired dynamic model and real-time control systems. 

To express how this IK method can be integrated with dynamic models and can be used for trajectory tracking in experimental setups, a model-based control algorithm is required to assist with controlling the robot on a designed trajectory. 
Different dynamic models have been presented for this reason \cite{kriegman2020scalable,li2017kinematic, katzschmann2019dynamic, chirikjian1994modal, bruder2019modeling, godage2015modal, boyer2020dynamics}.
In this research, as the IK method is presented for the PCC model, a model-based control based on this modeling will be chosen.
Also, an experimental setup with a soft robot control unit has been used to show the real-time applications of the models presented in this study. 

To summarize, the contributions of this work include the following:
\begin{enumerate}
\item  A new fast and accurate IK model for trajectory tracking of extensible soft robots that is easy to use and efficient for multi-segment soft robots. 
\item A new approach towards tip-angle control of soft robots on desired trajectories.
\item An overview of implementing the IK model for open-loop and real-time closed-loop control  via extensive simulations and experiments.
\end{enumerate}

The organization of this work is as follow. After this Introduction, in section (\ref{sec:Material}) an overview of the IK method and also a description of the dynamic model and the proposed controller are presented. Then, an overview of the experimental setup which is used for validation of the models is discussed. Conclusively, in section (\ref{sec:resutls}) different numerical simulation are studied and results of the experiments are presented, followed by a brief conclusion.

\section{Material and Methods}
\label{sec:Material}

\subsection{Inverse Kinematics method}
\label{sec:IK_method}

\begin{figure}
    \centering
    \includegraphics[width=0.5\textwidth]{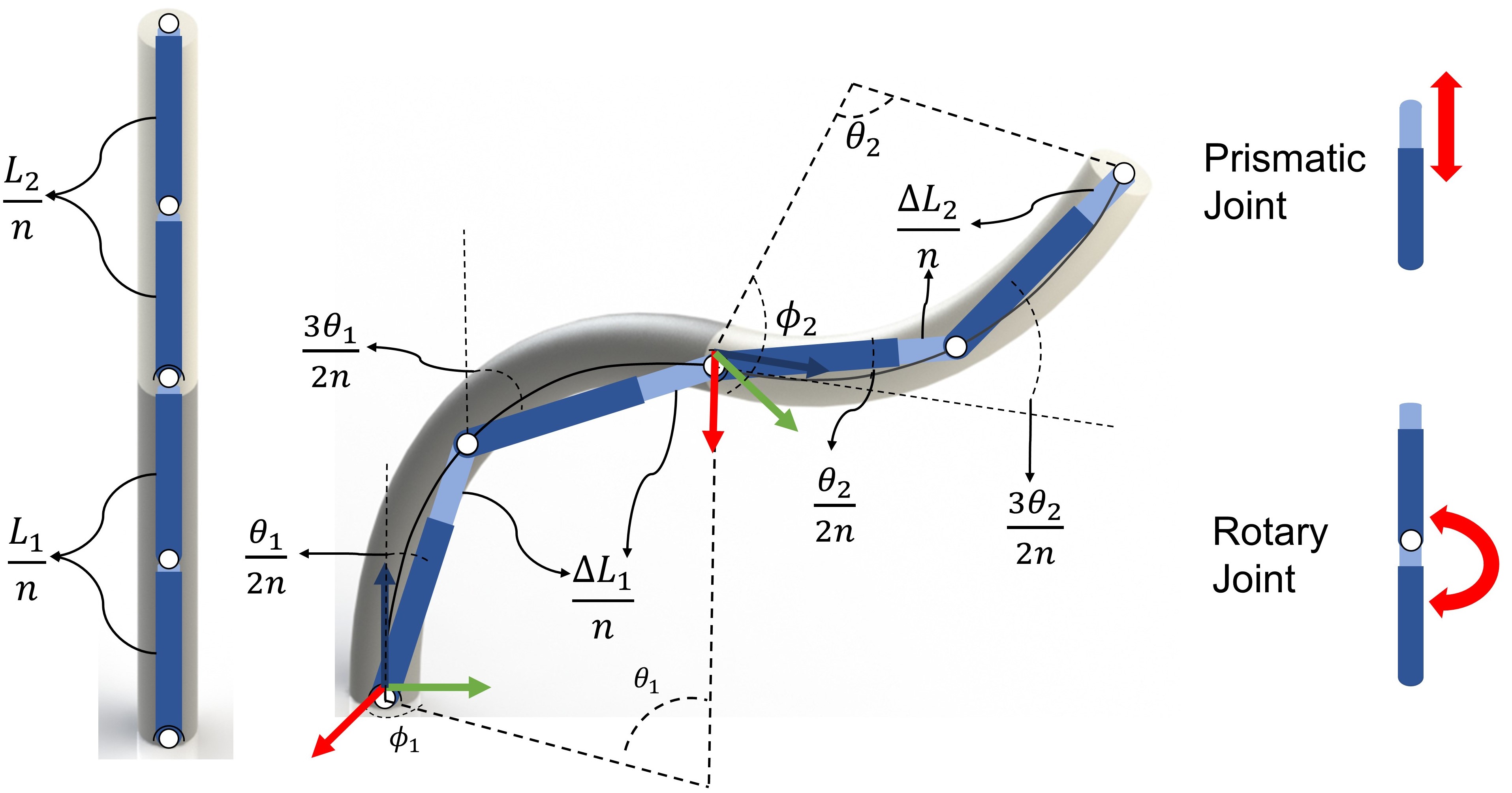}
   \caption{Combination of PCC and rigid-link model for the multi-segment soft robot and the respective configuration variables. The configuration before (left) and after actuation (right) is shown. Each PCC curvature is represented with two rigid links presented in dark blue connected to each other with rotary (circles), and prismatic (light blue) joints. Here, n is the number of links in each segment, L is the initial length of the segment, $\Delta L$ is the extension of each segment, $\theta$ is the bending angle, and $\phi$ is the deflection angle from X-axis.} 
    \label{fig:PCC_rigid}
\end{figure}

The complex dynamic behavior and geometries of soft robots make solving inverse kinematics more challenging than traditional rigid robots. One approach to reducing the complexity of soft robots is to discretize the robot curvature with small rigid links. This model is known as the rigid-link model and is mostly used in modeling human body motion and compliant mechanisms (\cite{Rigid_link_bio}\cite{mechanism_rigid_link}). The rigid links here are inextensible and connected via prismatic and rotary joints to cover both extension and rotation in dynamic behavior (Fig\ref{fig:PCC_rigid}). Applying this method to the soft robots makes it possible to implement previously developed methods for rigid robotics into modeling soft robots.


Moreover, since in many soft robots, the shape is uniform without torsional motion, the constant curvature approximation is valid. Therefore, to further simplify the dynamic behavior, the robot can be divided into multiple segments using PCC approximation and combined with the rigid-link model for a more straightforward approximation. In this approach, each robot segment is represented by multiple tiny rigid links connected to each other, forming part of a regular polygon. For extensible robots, the joints between the links are both rotary joints to emulate the bending of the actuator and prismatic joints to demonstrate the change in length of the robot (Fig.\ref{fig:PCC_rigid}). One primary assumption is that all rotary and prismatic joints move uniformly for each segment. Hence, the overall change in the length of each segment would be the deformation of each prismatic joint multiplied by the number of rigid links, and the same applies to the curvature. 

Although replacing the curvatures with multiple links reduces the complexity of the kinematic equation, it is not possible for the multi-segment robots to solve the problem analytically since the number of variables exceeds the number of kinematic equations. Thus, a numerical solution can be employed. From various numerical methods, we decided to use the optimization approach. In this method, the forward kinematics of the 3D robot is derived from the PCC and rigid-link model (Eq.(\ref{eq:multi_3D_ff})).

\begin{equation}
    \begin{aligned}
    \begin{bmatrix}
    X_{E_i}\\
    Y_{E_i}\\
    Z_{E_i}\\
    \end{bmatrix}  =  \begin{bmatrix}
    X_{E_{i-1}}\\
    Y_{E_{i-1}}\\
    Z_{E_{i-1}}\\
    \end{bmatrix} + \frac{(L_i+\Delta L_i)}{n}R_z (\phi_{i-1})R_y(\theta_{i-1})\\
    \begin{bmatrix}
    cos(\phi_i)\sum_{j=1}^{n}sin(\frac{2j-1}{2n}\theta_i)\\
    sin(\phi_i)\sum_{j=1}^{n}sin(\frac{2j-1}{2n}\theta_i)\\
    \sum_{j=1}^{n}cos(\frac{2j-1}{2n}\theta_i)\\
    \end{bmatrix} 
    \end{aligned}
    \label{eq:multi_3D_ff}
\end{equation}
Where $X_{E_i}$, $Y_{E_i}$, and $Z_{E_i}$ are the end-effector positions of the $i^{th}$ segment, $L_i$ is the length of the segment, $\Delta L_i$ is the change of length of the segment, $\theta_i$ is the bending angle of the segment, $\phi_i$ is the deflection angle from the X-axis, $R_y$ and $R_z$ are the rotation matrix around Y-axis and Z-axis respectively, and $n$ is the number of rigid links in each segment. It is worth mentioning that, although the IK method is presented for a 3D case, Eq.(\ref{eq:multi_3D_ff}) is also applicable for 2D systems. The only difference is the deflection angle ($\phi$) being zero in the 2D system.  After obtaining the forward kinematics, an optimization method is applied to find parameters that minimize the error between the desired point,$[X_d\,\,Y_d\,\,Z_d]$, and the end-effector point derived from the forward kinematics, [$X_{E_f}$\,\,$Y_{E_f}$\,\, $Z_{E_f}$], as in Eq.(\ref{eq:optimization}). The optimization approach is not limited to just the PCC model and can be applied to other soft robotics models.

\begin{equation}
\begin{aligned}
\min_{\Delta L,\theta,\phi} \quad & \lVert \begin{pmatrix} X_{d}-X_{E_f}, Y_{d}-Y_{E_f}, Z_{d}-Z_{E_f} \end{pmatrix} \rVert_2\\
\textrm{s.t.} \quad & \Delta L_{i_{min}}<\Delta L_i < \Delta L_{i_{max}}\\
  &   \theta_{i_{min}} < \theta_{i} < \theta_{i_{max}} \\
  &   -\pi<\phi_i<\pi
\end{aligned}
\label{eq:optimization}
\end{equation}


The optimization method applied here is MATLAB fmincon function which uses the Sequential Quadratic Programming algorithm(SQP). Since the SQP algorithm can not find the global optimized solution, the initial conditions and constraints are essential factors in finding the most optimal solution for the inverse kinematics. We set the initial conditions for optimization as the configuration variables of the robots at the previous posture to have a more smooth and energy-efficient solution over the trajectory and avoid abrupt changes in robot posture. In addition, since robots do not have unlimited changes in length or bend, physical constraints are added to the optimization equation to have a more realistic solution according to the robots' capability. Eq.(\ref{eq:optimization}) includes these physical constraints, which are the extension and bending limitations of the robot. Since the method is designed, so the configuration variables have minimum change between different points of the trajectory, the change of energy throughout the trajectory and hence, the overall energy consumption is minimum.

In the case of a multi-segment robot, since the configuration variables are more than the kinematic equations, there are multiple possible solutions for each desired position. These multiple solutions give the robot redundancy, which is beneficial for a secondary task such as tip angle control. The secondary task can be included in the inverse kinematics solution by adding additional constraints. This additional constraint for the tip angle control of the 2D robot is presented in Eq.(\ref{eq:tip_angle}).
\begin{equation}
    \theta_d - \sum_{i}^{n}\theta_i = 0
    \label{eq:tip_angle}
\end{equation}
Where $\theta_d$ is the desired angle and the $\theta_i$ is the bending angle of each soft segment. This additional constraint, the IK model benefits from the redundancy to not only find parameters that follow the trajectory but also calculate the bending angle of each segment in a way that keeps the tip angle at a certain angle. However, the priority of the IK model is to follow the trajectory and, if possible, achieve the second task. 
Using the boundary condition of the method enables us to add the secondary task. However, this method limits the IK model to satisfy the secondary task. Therefore, we manipulate the program in a way that the inverse kinematics solution is the priority, and if possible, it follows the second task, which is tip angle control. Furthermore, for other secondary tasks such as pressure optimization, or obstacle avoidance, using the boundary conditions is still applicable since we need both inverse kinematics and obstacle avoidance and pressure optimization to happen simultaneously. To the best of the authors' knowledge, adding second-priority tasks to the IK solution of soft robots has yet to be studied in previous studies.


\subsubsection{Error in Inverse kinematics method}
\label{error_IK}

Despite the method being straightforward to implement, one of its main limitations is the calculated length of the robot. As each curvature is modeled with multiple straight lines, the calculated length is smaller than the actual value. The first solution to overcome the drift is increasing the number of rigid links so that the error between the actual and the calculated length limits toward zero. However, the computation would be more expensive with this method, and calculation time would increase significantly.
In addition, in very high DoFs robots, this method may even affect the optimization solution and make the robot's motion unstable over a trajectory. The other solution to overcome this drift is a geometrical approach. The drift can be compensated by using the cosine law and geometrical equations to find the ratio between the actual length and the calculated length (Eq\ref{eq:cosine_law}). This equation is implemented at the end of each iteration of the inverse kinematics solution to compensate for the actual and calculated value difference.
\begin{equation}
    \frac{L}{\sum_{j=1}^{n}L^{'}} = \frac{\theta}{n \sqrt{2-2cos{\frac{\theta}{n}}}}
    \label{eq:cosine_law}
\end{equation}

Where $L^{'}$ is the length of each segment, $L$ is the length of the curvature, $n$ is the number of segments, and $\theta$ is the bending angle. For example, using Eq\ref{eq:cosine_law}, for a $90^\circ$ of bending angle, at least ten links are required to have an error length of $0.1\%$. However, having this many links in the IK model, especially for multi-segment robots,  significantly increases the computational costs. Therefore, as shown in \ref{sec:IK_result}, using Eq\ref{eq:cosine_law} helps to have an accurate solution even with just a few links representing the curvature.

\subsection{Dynamics and Control}
\label{sec:dynamics_control}
For the IK model to be studied, different trajectories will be applied to a three-bellowed soft robot, 
 this will show the integration of the IK model with dynamic models. A model-based controller is employed to control the robot on the desired trajectory. In this section, the dynamics and also the controller applied to the robot are presented. 

For the dynamics of the PCC model, a bishop frame is attached to every point on the backbone curve (in SE(3)), which is parameterized by a state vector $q = [\epsilon\,\,k_x\,\,k_y]^T \in R^3 $. Here, $\epsilon$ is the strain of the robot 
, $k_x$ and $k_y $ are the curvatures of the robot in the x-z and y-z plane, respectively. The relation between the newly defined state variables and the configuration variables defined in the IK model (Fig.(\ref{fig:PCC_rigid})) is as follows:
\begin{subequations}
  \label{eq:defin(dof)}
\begin{equation}
  \label{eq:q1}
  \begin{aligned}
  \epsilon=(l-l_0)/l_0 \\ 
\end{aligned}
\end{equation}
\begin{equation}
  \label{eq:q2}
  \begin{aligned}
 k_x=cos(\phi)\times\theta/l\\
\end{aligned}
\end{equation}
\begin{equation}
  \label{eq:q3}
  \begin{aligned}
   k_y=sin(\phi)\times\theta/l\\
\end{aligned}
\end{equation}
\end{subequations}
\noindent
The position of each point on the backbone curve is defined as:
\begin{equation}
 \label{eq:backbone_position}
  \begin{aligned}
p(q,\sigma)= \int_{0}^{\sigma} \Phi(q , \eta)U(q)d\eta 
\end{aligned}
\end{equation}
\noindent
Where $\Phi (q,\eta) $ and $U $ describe the differential geometry of the backbone curve. Thus, the equation of motion of the soft robot can be expressed as:
\begin{equation}
\label{eq:eq_of_motion}
  \begin{aligned}
M(q)\ddot{q}+C(q, \dot{q}) \dot{q}+G(q)+K(q)+D(\dot{q})=\tau
\end{aligned}
\end{equation}
\noindent
Where, $M(q) \in {\rm I\!R}^{n\times n}$ is the inertia matrix, $C(q,\dot{q})\dot{q} \in {\rm I\!R}^{n}$ contains the Coriolis and centrifugal forces, and $G(q) \in {\rm I\!R}^{n}$ is the gravitational forces acting on the robot.
\noindent
As mentioned previously, there exists a rich literature on dynamic modeling of PCC models \cite{Katzschmann1, Cossimo_journal1,Marchese}. In this study, a dynamic model is needed that can be used for real-time control of multi-link soft robots and can perform efficiently and accurately. So, the model presented in \cite{Brandon_journal1} is employed in this study. 
In Eq.(\ref{eq:eq_of_motion}), two matrices of $K(q)$ and $D(\dot{q})$ define the hyper-elastic and visco-elastic properties of the robot. The hyper-elastic potential energy is defined as:
\begin{equation}
 \label{eq:potentail_energy}
  \begin{aligned}
U_e(q) = \int_{}^{} k_e(\eta)\eta d\eta + \int_{}^{} k_b(\eta) \eta d\eta, \\
\end{aligned}
\end{equation}
\noindent
where $k_e(\eta)$ and $k_b(\eta)$ are elongation and bending stiffness. 
The elongation stiffness is defined as $k_e(1)=\alpha_1 +\alpha_2(tanh[\alpha_3 \epsilon]^2 -1)$.
Also, due to the layout of the pneumatic bellows of the soft actuator, the bending stiffness is asymmetric and defined as $k_b(1)=\alpha_\phi(q).[\alpha_4+\alpha_5(tanh[\alpha_6\beta]^2 -1)]$,  where, $\alpha_\phi(q)= 1/2\beta [sin(m.\phi)+1]+1$.
The overall stiffness matrix is presented  as: 
\begin{equation}
 \label{eq:K}
  \begin{aligned}
K(q)=\frac{\text{d}U_e}{\text{d}q}\\
\end{aligned}
\end{equation}
\noindent
Also, the Rayleigh damping matrix is defined as $R$, and:
\begin{equation}
 \label{eq:D}
  \begin{aligned}
D(\dot{q})=R \dot{q}\\
\end{aligned}
\end{equation}
\noindent
 In this modeling, the generalized input vector is chosen to be  $\tau(u)= Hu$ with $H$ being a mapping from input space to joint actuation space. Using \cite{Brandon_journal1}, $H$ is chosen as in Eq.(\ref{eq:H}), where $\gamma_i= (i-1).2\pi/m $ and $m$ is the number of pneumatic bellows. Here $\alpha_1>0$ and $\alpha_2>0$ have to be identified using experimental data.
 \begin{equation}
 \label{eq:H}
  \begin{aligned}
H= \begin{bmatrix}\alpha_1 & \alpha_1&\alpha_1   \\- \alpha_2cos(\gamma_1)& - \alpha_2cos(\gamma_2) &- \alpha_2cos(\gamma_3)\\ \alpha_2cos(\gamma_1)&  \alpha_2cos(\gamma_2) & \alpha_2cos(\gamma_3)\\ \end{bmatrix}\\
\end{aligned}
\end{equation}
\noindent

The aim of this research is to control the soft manipulator to follow a desired trajectory in the Cartesian space. Using the inverse kinematics, the equivalent goal would be for the robot to reach a desired posture in the state space, which theoretically means $\lim_{t \to \infty}q=q_d$. It is worth emphasizing that since the IK method is independent of the controller, any controller that works in the state space can be used with this IK method, which gives this method a priority over learning methods.
As mentioned previously in Eq.(\ref{eq:eq_of_motion}), the two matrices of $K(q)$ and $D(\dot q)$ exhibit the hyper-elastic and visco-elastic properties of the system. Due to this property of soft robots, as discussed in \cite{della2021model}, soft robots have a self-stabilizing feature that proves to be advantageous for controlling purposes. 
With this analysis, the controller used for trajectory tracking is:
\begin{equation}
 \label{eq:controller1}
  \begin{aligned}
\tau&=\ddot{q}+K(q_d)+G(q_d)\\
where, \ddot{q}&=\ddot{q_d}-k_v(\dot{q_d}-\dot{q})-k_p(q_d-q)\\
\end{aligned}
\end{equation}
\noindent
Where two terms of $K(q_d)$, the stiffness matrix, and $G(q_d)$, the gravity matrix, are the feedforward terms and $\ddot{q}$ is chosen, so the feedback controller is a PD controller. 
Here, $q_d$ is the desired trajectory in the state space which has been designed using the inverse kinematics model. 
Also, $\ddot{q_d}=0$, and as the trajectory tracking is assumed to be slow, $\dot{q_d}=0$. The graphical interpretation of this controller is pictured in Fig.(\ref{fig:Intro}).


\begin{figure*}[h!]
\centering
\includegraphics[width=\textwidth]{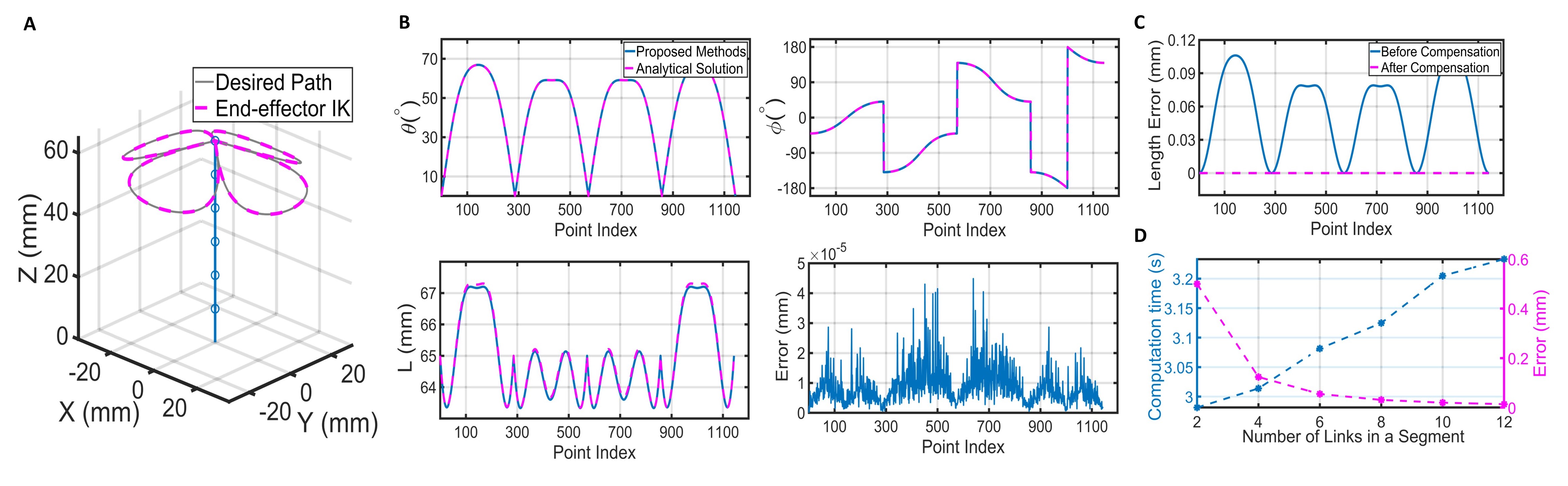}
\caption{The inverse kinematics model results for a single-segment robot over a given trajectory divided into 1000 points. (A) the robot configuration and calculated end-effector position using IK model (B), The calculated parameters at different points of the trajectory ($\theta$, $\phi$, $L$) using the IK model and analytical solution, and the error between the desired position and the IK model over the trajectory. (C) The difference between the length before and after compensate for the drift using Eq\ref{eq:cosine_law}. (D) The computation cost of solving the inverse kinematics versus the number of links in each segment. By decreasing the number of links, the computation cost decreases, but the length error increases significantly.}
\label{single_seg_3d_ik}
\end{figure*}

\section{Results and Discussions}
\label{sec:resutls}
Using the proposed methods in sec.(\ref{sec:Material}), a set of simulations and experiments that can demonstrate how a soft manipulator can use the IK model to track a trajectory is presented. 

\subsection{Single-Segment Soft Robot}
\label{sec:IK_result}
In the first step, the proposed inverse kinematics solution is applied to a single-segment 3D soft robot that follows a four-sided flower trajectory. Since an analytical solution is possible in this case, the obtained results can be compared with analytical parameters for validation. Here, the analytical solution is from \cite{Analytical}.
As presented in Fig.(\ref{single_seg_3d_ik}-A), the IK model follows the desired trajectory with an average error of $8.41\times 10^{-6} mm$ which is less than 0.001$\%$ of the workspace. It can also be detected that the calculated bending and deflection angles are identical for both analytical and presented IK models. The only difference between these two solutions is the length of the robot. This difference was expected as the actual curvature is longer than the rigid-link approximation. As explained in section\ref{error_IK}, there are two solutions to overcome this error, either increasing the number of links in each segment or applying geometry compensation. However, as presented in Fig.(\ref{single_seg_3d_ik}-D), increasing the number of links in each segment increases the computation cost significantly. Therefore, in order to compensate for this drift, a coefficient dependent on the angle and the number of segments is multiplied by the calculated length. 
The computation cost for six links and 1150 points over the given trajectory is 3.07 seconds. In addition, the average computation cost for one point on the trajectory is just 0.28 ms which is less than reported for kinematics in similar studies such as \cite{computation_cost}. As shown in Fig.(\ref{single_seg_3d_ik}-C), the coefficient removes the drift between the actual and calculated length. With this drift compensation, the parameters obtained from the IK model are identical to those from the analytical solution, showing how accurately the model tracks a trajectory.



\subsection{Multi-Segment Soft Robot}
\label{sec:IK_result_2}

\begin{figure*}[h!]
\centering
\includegraphics[width=\textwidth]{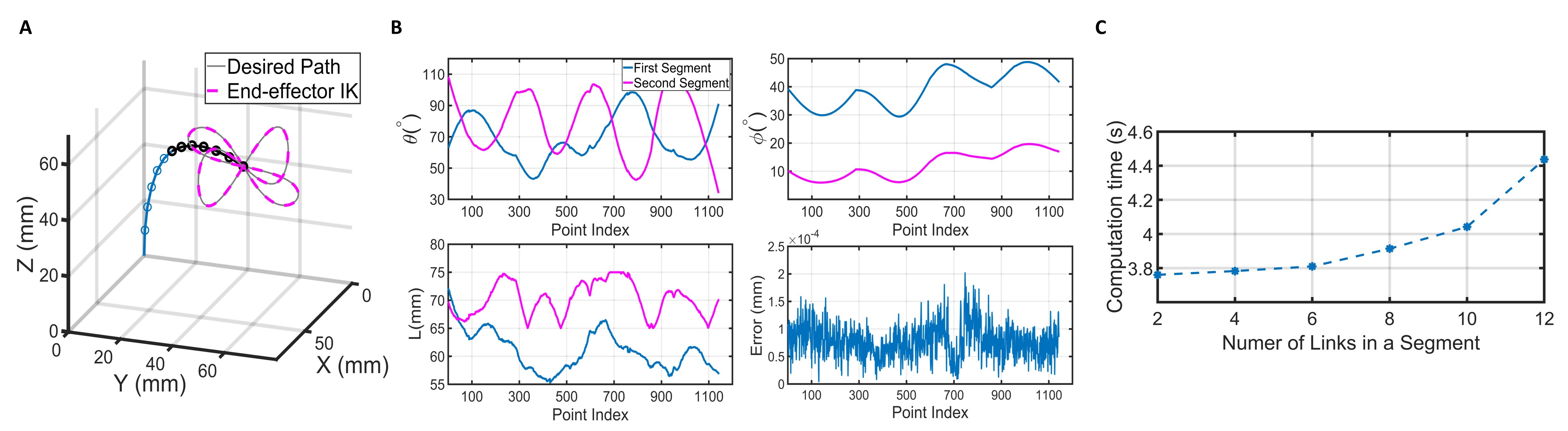}
\caption{The inverse kinematics model results for a two-segmented robot, the first segment is shown in blue and second segment in black, (A) the robot configuration and calculated end-effector position using IK model, (B) The calculated parameters ($\theta$, $\phi$, $L$) of both segments using IK model, and the error between the desired position and the IK model solution. The average error in this case is $1.95\times10^{-5} \%$ of the workspace. (C) The computation time of the inverse kinematics versus the number of links in each segment.}
\label{multi_seg_3d_ik}
\end{figure*}

Unlike for the single-segment robot, the analytical solution is not possible for multi-segment soft robots since the number of DoFs exceeds the number of kinematic equations; thus, the calculated parameters can not be compared with a particular result. Here, we study a two-segment soft robot that is going to track a 3D trajectory as shown in Fig.(\ref{multi_seg_3d_ik}).
As can be seen, the average error between the end-effector and the desired position over the given trajectory is $7.78\times10^{-5}mm$, and this is negligible and lower than the reported error in similar prior work,  \cite{Jacobian_learning},\cite{underwater}. In addition, as shown in the calculated parameters, Fig.(\ref{multi_seg_3d_ik}-B), there are no abrupt changes in the change of length, the bending angles, or the deflection angle, which leads to the smooth motion of the robot over the given trajectory.

\begin{figure}[h!]
\centering
\includegraphics[width=0.48\textwidth]{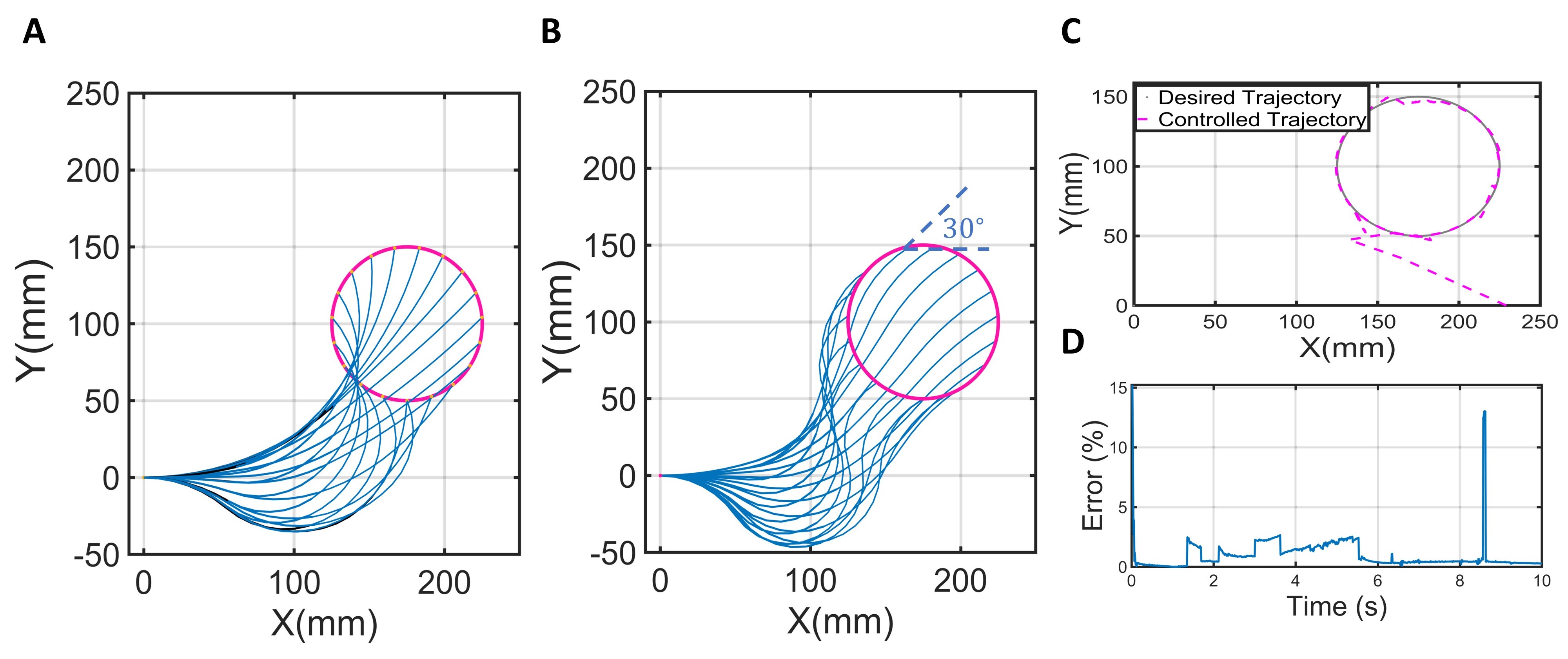}
\caption{Comparison of two different IK solutions on the same trajectory for a four segment planar soft robot, (A) the robot only has the task of following the desired trajectory (B) the robot has a second task of keeping the tip angle a constant value of $30^o$ with respect to the horizon. (C) simulation results of the robot being controlled on the desired trajectory pictured in (B), grey line is the desired trajectory and the pink dashed line is the trajectory followed by the robot. Initial state of the robot is $q_0=[0,0,0,0,0,0,0,0]$, (D) the norm of error between the desired trajectory and the position of the end-effector}
\label{fig:IK_4link_circle_control}
\end{figure}

Using the additional constraint defined in Eq.(\ref{eq:tip_angle}), it is also possible to design trajectories for a robot with a constant tip angle. To further study this, a 4-segment 2D soft robot is chosen to follow the desired trajectory of  $X_d=175+50cos(t)$ and $Y_d=100+50sin(t)$. This trajectory is given as input to the IK code. The IK method is employed without the additional tip angle constraint in Fig.(\ref{fig:IK_4link_circle_control}-A). It can be seen that the state variables are calculated at every step so that the change of these variables with respect to the previous step on the trajectory is minimum, and hence, energy consumption is minimum. This is a key factor in generating smooth trajectories without unwanted jumps in the robot's state variables. Since, in this example, the robot has redundancy with respect to the task space, except for following the desired trajectory, it is possible to add tip angle constraints on the robot with lower priority. As explained in section(\ref{sec:IK_method}), an extra constraint is added to the robot's trajectory with the aim of having a constant tip angle throughout the trajectory tracking. Fig.(\ref{fig:IK_4link_circle_control}-B) depicts how different the robot moves as it is constrained to keep the tip angle constant. 

It has been shown that the IK model is able to generate smooth 2D and 3D results even on complex trajectories. The next step is to demonstrate how it can be used as an input to dynamic models. 
To be able to control the robot on desired trajectories, the physical and material parameters of the robot need to be identified. The soft manipulator used in this research 
has the following physical parameters: the mass $m_0=17.3 g$, and the relaxation length of $l_0=64.4 mm$. 
Considering preliminary uni-axial tension tests, the 3D-printed elastomer material is estimated to be linear isotropic with Young's modulus of $E = 80$ MPa and a Poisson ration $\nu = 0.49$. Considering the geometry and material of the soft robot, the hyper-elastic and visco-elastic material parameters and the Rayleigh damping matrix are chosen similarly to \cite{Brandon_journal1} (their Table 1), and no further identification is required.

For the first simulation, the circular trajectory defined as the desired trajectory for the 4-segment 2D soft robot is chosen (Fig.(\ref{fig:IK_4link_circle_control})). 
Here, using the IK model, the desired state variables for each robot segment are calculated, and using the model-based controller, the robot is controlled on the desired trajectory.
To control the robot, a combination of feedforward and feedback control, as explained in section(\ref{sec:dynamics_control}) has been employed. For the sake of brevity, only the results of the controller applied to only the trajectory of the Fig.(\ref{fig:IK_4link_circle_control}-B) have been reported in Fig.(\ref{fig:IK_4link_circle_control}-C and D).  The PD controller used in this case is formulated as $\tau= B^{-1}[M(\ddot{q}-k_v(\dot{q_d}-\dot{q})-k_p(q_d-q))+K(q_d)+G(q_d)]$ with $k_p=1000\times I_8$ with $k_v= 5 \times I_8$. 
The robot starts tracking the trajectory from the initial state of zero, and as in Fig.(\ref{fig:IK_4link_circle_control}-D), the norm of error of following this trajectory is less than $2.5 \% $ which is acceptable. The results presented in Fig.(\ref{fig:IK_4link_circle_control}) can verify that the results of IK are very smooth and even controlling a multi-segment soft robot on a desired trajectory with a second desired task does not have any abrupt changes. 
This is an advantage of the presented method compared to some existing ones, such as \cite{Jacobian_learning}.
Also, the controller used in this example is independent of the IK model, which is an advantage to the proposed IK model.  

\subsection{Experiments}
\label{sec:experiments}
An experimental setup, including a soft robot and a hardware system to control the robot, is demanded to show how the IK model and designed controllers can be implemented on a soft robot. To this end, a 3D printed soft manipulator consisting of 3 parallel embedded pneumatic bellows is used; this robot is pictured in Fig(\ref{fig:Intro}). By inflating or deflating each bellow, the manipulator can change its posture. If all the bellows are actuated simultaneously, the robot changes length, and with unequal pressurization of the bellows, the robot can create a pose that forms a curve with constant curvature. Hence, this robotic system is suitable for experiments in this research case, as it can represent a PCC model. Furthermore, since this type of robot has already been used in \cite{Brandon_journal1}, the dynamic model presented in section(\ref{sec:dynamics_control}) is also compatible with this experimental setup.


A Soft Robotic Control unit (SRC), presented in \cite{caasenbrood2022desktop}, has been used to create fast, accurate model-based control. In this control unit, the control and data acquisition is done using a Raspberry Pi 4 (2GB RAM). A proportional piezo-actuated pressure regulator (Festo, VEAB-L-26-D7-Q4-V1-1R1) with a custom Raspberry Pi Hat simultaneously allows pressure measurements. The main software of this setup 
is designed to read multiple sensors, simultaneously control several VEAB regulators, and communicate with other devices via TCP/IP.
It is important to briefly emphasize that due to large continuous deformations and potential changes in dynamics, sensing and adding sensors is known to be quite challenging in soft robotics. For this reason, to record the angular deflections of the soft robot, a 9-DOF inertial measurement unit (Bosch, BNO055) is attached to the robot's end-effector. The information from the IMU sensors is read directly from the SRC software and can be used throughout the real-time control of the soft robot. In this control setup, the sensor is connected via $I^2 C$ to the SRC. 
In order to read the Cartesian movements of different points on the backbone curve of the soft robot, an OpenMV H7 R2 Camera is employed. Using colored markers and color detection methods, the position of the markers on the backbone is recorded, and with that, the length of the robot is determined. Hence, all state variables of the robot can be known throughout the experiments. In this setup, the regulators have a 100-150 ms delay, and the control rate can reach a max of 300-400 Hz. 


To start the experiments, first, the parameters for the mapping matrix defined in Eq.(\ref{eq:H}) are identified. For this reason, different experiments are run, and in each experiment, a set of quasi-static pressure is added to the robot. With each set of pressure, the robot reaches a final point in the cartesian space. This point can be identified using the data from the OpenMv camera ($[X,Y,Z]_{markers}$) and IMU sensors ($[roll, pitch, yaw]$). So one can identify the desired state variable of the dynamic model. These state variables are fed into the model, and the required input pressure for the robot is calculated via simulations, which are then compared with the pressure input of the experiments. With this comparison, the parameters are identified for the mapping matrix of Eq.(\ref{eq:H}), as $\alpha_1 =8\times 10^{-4} $  and $\alpha_2= 7.91\times 10^{-7} $. 
It is worth mentioning that since in these experiments we are using the IMU sensors, only one position data from the cameras is sufficient to find the state variable, here we use the $Z_{markers}$. 

\begin{figure}[t]
\centering
\includegraphics[width=0.49\textwidth]{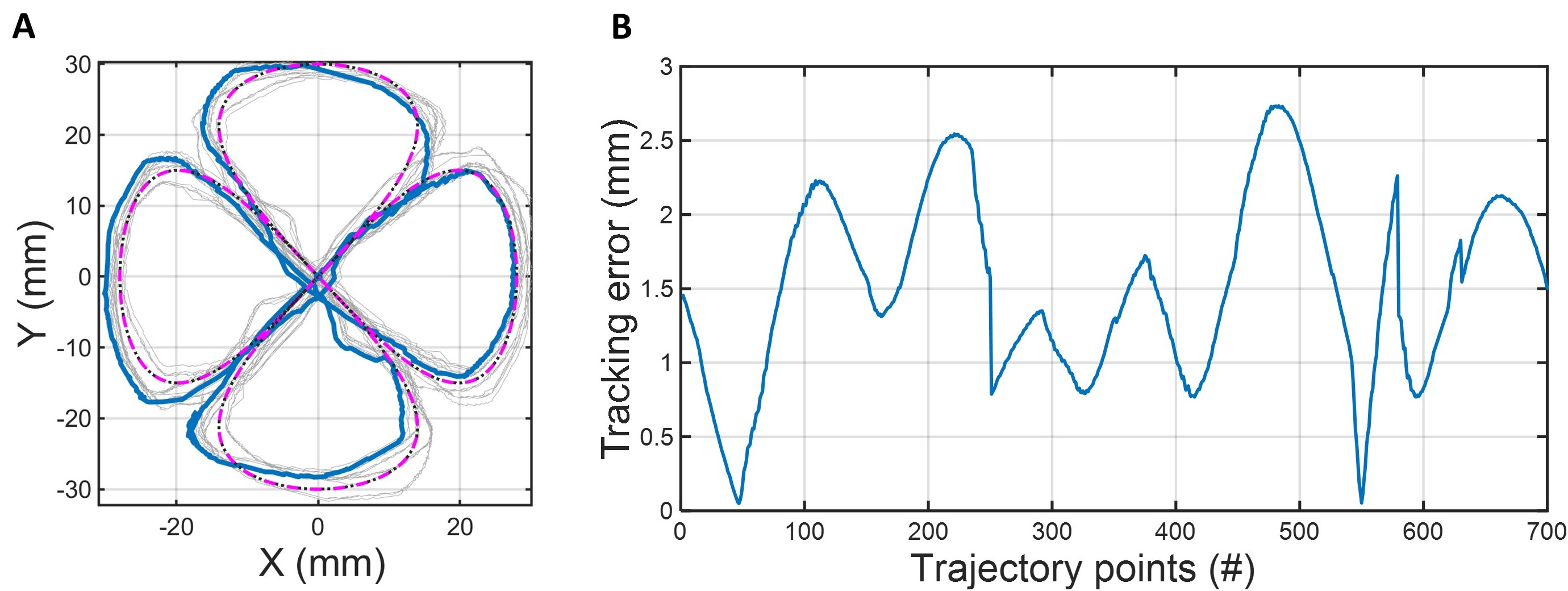}
\caption{Experiment 1 - (A) Top view of the control of the robot on the designed trajectory of Fig.(4). Dashed pink plot is the desired trajectory, the grey plots are different experiments, and the continuous blue plot is the average of the followed trajectory by the robot. (B) The norm of the error between the followed trajectory by the tip point and the desired one. Video of the experiment available at: \url{https://youtu.be/1kQ8FbXpwf0}}
\label{fig:open_loop_control}
\end{figure}

Having a bridge between the model and the experimental setup, different trajectories in experiments are achievable, and this can be used for further validation of the model. First, an open loop controller is employed to control the robot on the desired trajectory presented in Fig.(\ref{single_seg_3d_ik}). Here, the trajectory is added to the dynamic model, and the desired pressures calculated from the simulations of the closed-loop controllers are added to the robot to perform an open-loop control. The top view of the final trajectory tracking of the robot is presented in Fig.(\ref{fig:open_loop_control}). The error between the desired trajectory and the following one is less than 3 mm, which is acceptable considering the trajectory's complexity for a three-bellowed robot and the fact that we are using an open-loop controller. In this experiment, as shown in Fig.(\ref{single_seg_3d_ik}), the robot experiences a change of length, and this shows that the IK model is also valid for trajectories with the change of length.

In the second experiment, a closed-loop controller is designed to keep the robot's end-effector on the trajectory of $X_d=20cos(6t)$, $Y_d=20sin(6t)$, $Z_d=60$. 
For both experiments, the initial condition is zero, $q_0=[0,0,0]$, and the control parameters are chosen as $k_p=0.16 \times I_3$ and $k_v\approx 0\times I_3$.

\begin{figure*}[t]
\centering
\includegraphics[width=1\textwidth]{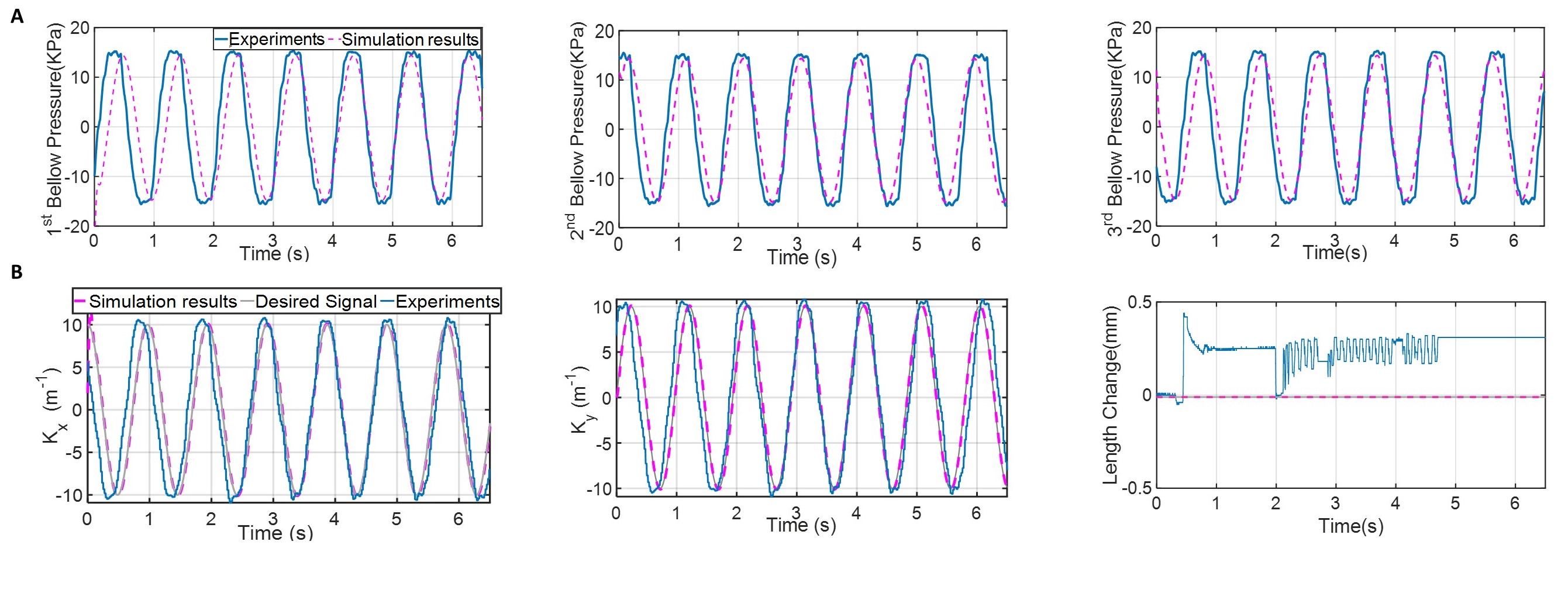}
\caption{Experiment 2 - (A) Pressure values, dashed lines are the designed values and blue lines are the regulator measurements (B) State variables of the robot in two different planes indicating state variables of the dynamic model. The grey plots are the results of IK model which are the desired trajectory, the dashed lines show the simulations results of the controller, and the blue lines show the movements of the robot in experiments.} 
\label{fig:experiment2}
\end{figure*}

In every step of the real-time control, first, the control of the robot is simulated using a feedback controller presented in Eq.(\ref{eq:controller1}). At this step, we have a set of three pressures that indicate how much each bellow needs to be pressurized to follow the desired trajectory. 
These values are fed directly into the robot using pressure regulators. As the robot is pressurized, the IMU sensors and camera data send feedback to the model to help calculate the error with respect to the desired state variables. From there, a new set of the pressure signal is calculated. The designed controller in this study is programmed in MATLAB/Simulink and communicates via TCP/IP with the SCR unit at 250 Hz.
As the experiments run, it can be verified that the robot is able to follow the desired trajectory. Fig.(\ref{fig:experiment2}-A) pictures the designed values for the pressure of each bellow and compares it with the pressure produced by the regulators in each bellow. It depicts how these two values overlap after the initial transient state is finished, meaning the experimental setup can generate the pressure inputs very close to the designed pressure values, which helps the trajectory tracking for the soft robot. 
Also, using the data from the camera and the IMU sensors, it is possible to study the state variables of the robot as it is tracking the trajectory. Fig(\ref{fig:experiment2}-B) compares these state variables between the designed values from the IK solution, the model simulations, and the controlled values in the experiments. With this figure, it is also verified that trajectory tracking can be achieved with the combination of the IK model, the dynamic model, and the experimental setup.

\section{Conclusions}
This research aims to propose an inverse kinematic model for soft extensible actuators that can be modeled with a piece-wise constant curvature approach, whether 2D or 3D, single-segment or multi-segment. In this method, each segment of the soft actuator was modeled with multi-rigid links that are connected through rotatory and prismatic joints. To have lower computational costs, we proposed a minimum amount for the number of rigid links without changing the accuracy of the results. The approach uses a simple procedure to minimize the error between the desired position of the end-effector and the forward kinematics equations. In this energy-efficient method, adding a second desired task, such as tip angle control, is possible using additional constraints. The presented method is verified through simulations and different experimental results, which have validated the method's applicability to different trajectories. 
For future work, we aim to focus on achieving a more comprehensive range of secondary tasks within the algorithm, including tasks such as obstacle avoidance. Also, we aim to use other optimization methods, such as a generic algorithm, to find the globally optimized points and have much higher precision.
One main advantage of the presented method is the independence of the controller applied to the robot. Regardless of the type of controller and the dynamic model, the IK model has a low computational cost and high accuracy for trajectory tracking.

\bibliographystyle{IEEEtrans}
\bibliography{reference} 

\end{document}